\documentclass[journal]{IEEEtran}

\usepackage[usenames]{color}
\usepackage[utf8]{inputenc}
\usepackage{graphicx}
\usepackage[section]{placeins} 
\usepackage{amsfonts}
\usepackage{amssymb}
\usepackage{amsbsy}
\usepackage{amsmath}
\usepackage{multirow}
\usepackage{array}
\usepackage{color}
\usepackage{soul}
\usepackage{cite}
\usepackage[normalem]{ulem} 
\usepackage{times}
\usepackage[all]{xy}
\usepackage{lscape}
\usepackage{flushend}
\usepackage{url}
\usepackage{float}
\usepackage{rotating}
\usepackage{colortbl}

\usepackage[caption = false]{subfig}
\usepackage{xspace}

\graphicspath{{./images/}}
\DeclareGraphicsExtensions{.pdf,.png}

\def\Real{\mathbb{R}}

\def\K{\ensuremath{\mathbf{K}}}

\def\M{\ensuremath{\mathbf{M}}}

\def\X{\ensuremath{\mathbf{X}}}

\def\Z{\ensuremath{\mathbf{Z}}}

\def\x{\ensuremath{\mathbf{x}}}
\def\y{\ensuremath{\mathbf{y}}}
\def\z{\ensuremath{\mathbf{z}}}

\renewcommand{\baselinestretch}{0.99}

\usepackage{amsfonts}
\usepackage{amssymb}
\usepackage{amsbsy} 
\usepackage{amsmath}
\usepackage{graphicx}
\usepackage{epstopdf}
\usepackage{amsthm}
\usepackage{bbm}
\usepackage[table,xcdraw]{xcolor}
\usepackage[colorinlistoftodos,prependcaption]{todonotes}
\newcommand{\imag}{\mathbbm{i}}

\usepackage{array}
\newcolumntype{L}[1]{>{\raggedright\let\newline\\\arraybackslash\hspace{0pt}}m{#1}}
\newcolumntype{C}[1]{>{\centering\let\newline\\\arraybackslash\hspace{0pt}}m{#1}}
\newcolumntype{R}[1]{>{\raggedleft\let\newline\\\arraybackslash\hspace{0pt}}m{#1}}

\usepackage{pgffor} 
\usepackage{float} 
\usepackage{wrapfig}

\usepackage{amsmath}
\usepackage{amssymb}
\usepackage{amsthm}
\usepackage{algorithm}
\usepackage{algorithmic}
\usepackage{placeins}
\usepackage{multirow}
\usepackage{url}
\usepackage{cite}
\usepackage{wrapfig,lipsum,booktabs}
\usepackage[font=small,labelfont=bf]{caption}
\usepackage{graphicx}
\usepackage{xcolor}
\usepackage[all]{xy}
\usepackage{hyperref}
\usepackage{adjustbox}
\def\x{{\mathbf x}}
\def\z{{\mathbf z}}
\def\Z{{\mathbf Z}}
\def\X{{\mathbf X}}
\def\K{{\mathbf K}}
\def\R{{\mathbf R}}
\def\M{{\mathbf M}}

\def\bSigma{{\boldsymbol \Sigma}}

\def\Real{{\mathbbm{R}}}

\title{Efficient Nonlinear RX Anomaly Detectors}

\author{Jos\'e A. Padr\'on Hidalgo, Adri\'an P\'erez-Suay, Fatih Nar, and Gustau Camps-Valls
}

\date{}

\begin{document}

\maketitle

\begin{abstract}
Current anomaly detection algorithms are typically challenged by either accuracy or efficiency. More accurate nonlinear detectors are typically slow and not scalable. In this letter, we propose two families of techniques to improve the efficiency of the standard kernel Reed-Xiaoli (RX) method for anomaly detection by approximating the kernel function with either {\em data-independent} random Fourier features or {\em data-dependent} basis with the Nystr\"om approach. We compare all methods for both real multi- and hyperspectral images. We show that the proposed efficient methods have a lower computational cost and they perform similar (or outperform) the standard kernel RX algorithm thanks to their implicit regularization effect. Last but not least, the Nystr\"om approach has an improved power of detection.
\end{abstract}

\label{sec:intro}

\section{Introduction} 
Anomaly detection (AD), as a remote sensing (RS) research topic, is gaining importance because of the need for processing large number of images that are acquired from satellite and airborne platforms \cite{AD2}. AD aims to detect small portions of the image which do not belong to the background of the scene. Unlike target detection, AD does not use known target spectra and anomalies are assumed to be rare and at the tail of the background distribution. 

Among the many detector algorithms found in the literature, the Reed-Xiaoli (RX) detector~\cite{Reed90tassp} is widely used due to its good performance and simplicity.
The RX detector determines target pixels that are spectrally different from the image background based on the Mahalanobis metric. For RX to be effective, anomalous targets must be sufficiently small compared to background and is assumed to follow a Gaussian distribution.
However, it has been shown that the Gaussian distribution assumption fails, for example, in hyperspectral images or with complex feature relations, especially at the tails of the distribution\cite{Matteoli2010}. As a result, nonlinear versions of RX have been introduced to mitigate this problem, and the kernel RX (KRX) detector was proposed in~\cite{Heesung05tgars} to cope with complex and nonlinear backgrounds. However, the KRX algorithm has not been widely adopted in practice because, being a kernel method, the memory and computational cost increase, at least quadratically, with the number of pixels. This poses the perennial problem of accuracy versus usability in nonlinear detectors in general and kernel anomaly detectors in particular.

In this study, we focus on improving the space (memory) and time efficiency of the KRX anomaly detector. Kernel-based anomaly detectors provide excellent detection performance since they are able to characterize non-linear backgrounds \cite{CampsValls09wiley}.
In order to undertake this challenge, we propose to use efficient techniques based on random Fourier features and low-rank approximations to obtain improved performance of the KRX algorithm. 
We reported our initial efforts using the random Fourier features approach in\cite{RRX}

In the literature, {\em local} and {\em global} RX-based detectors have been proposed. 
In local AD~\cite{Reed90tassp}, pixels in a sliding window are used as input data. Despite their adaptation to local relations, the detection power has been shown to be low recently \cite{Guo2016, Matteoli2010}. 
Conversely, in global AD all image pixels are used to estimate statistics. Thereby, targets with various sizes and shapes can be detected while detection of small targets can be difficult.
In this study, all the methods are used in a global setting for the sake of simplicity.

\section{RX Based Anomaly Detection} 
\label{sec:rx_based_detection}
Among the various AD methods proposed in the literature, one of the most frequently used anomaly detectors is the Reed-Xiaoli (RX)~\cite{Reed90tassp}. In this section, we explain the RX method and its kernelized version, the KRX anomaly detector.

\subsection{RX Anomaly Detector}

We consider an acquired image reshaped in matrix form as ${\X}\in\mathbbm{R}^{n\times d}$, where $n$ is the number of pixels and $d$ is the total number of channels acquired by the sensor. For simplicity, let us assume that ${\X}$ is a centered data matrix.
The RX detector characterizes the background in terms of the covariance matrix $\bSigma = \frac{1}{d}{\X}^\top {\X}$. The detector calculates the squared Mahalanobis distance between a test pixel ${\x}_\ast$ and the background as follows:
\begin{equation}
\label{eq:rx_centered}
    D_{RX}({\x}_\ast) = {{\x}^{\top}_\ast} \bSigma^{-1} {{\x}_\ast}.
\end{equation} 
In a global AD setting, as discussed here, $\bSigma^{-1}$ can be efficiently computed using all the image pixels since the dimensionality of the image is much lower than the number of pixels ($d \ll n$).
Whereas, in a local AD setting, $\bSigma_p^{-1}$ needs to be computed for each image pixel $p$ using the centered pixels in a window having an origin at that pixel~\cite{Matteoli2010}.

\subsection{KRX Anomaly Detector}
It is known that linear RX is computationally efficient and leads to an optimal solution when pixels in $\X$ follow a Gaussian distribution. 
However, real life problems are not always Gaussian distributed and this requires more flexible models. 
Kernel methods are a possible solution because they can capture higher-order (nonlinear) feature relations, while still using linear algebra operations~\cite{CampsValls09wiley}.

In order to develop the kernel RX, let us consider a mapping for the pixels in the image to a higher dimensional Hilbert feature space ${\mathcal H}$ by means of the feature map $\boldsymbol{\phi}: \x\in\Real^d\to\boldsymbol{\phi}(\x)\in\Real^{d_\mathcal H}$. 
The mapped data matrix $\X\in\Real^{n\times d}$ is now denoted as $\boldsymbol{\Phi}\in\Real^{n \times d_{\mathcal H}}$. 
Let us define a kernel function $K$ that, by virtue of the Riesz theorem, can evaluate (reproduce) the dot product between samples in ${\mathcal H}$, i.e. $K(\x,\x')=\langle\boldsymbol{\phi}(\x),\boldsymbol{\phi}(\x')\rangle\in{\mathbb R}$. 

To estimate how anomalous a pixel is using a pixel under test for $\x_\ast\in\Real^d$, we first map it $\boldsymbol{\phi}(\x_\ast)$, and apply the RX formula in~\eqref{eq:rx_centered} as 
\begin{equation}
\label{eq:krx_Hilbert}
D_{KRX}(\x_\ast) = {\boldsymbol{\phi}(\x_\ast)}^\top(\boldsymbol{\Phi}^\top\boldsymbol{\Phi})^{-1}\boldsymbol{\phi}(\x_\ast)
\end{equation} 
which, after some linear algebra, can be expressed in terms of kernel matrices~\cite{Kwon07,CampsValls09wiley}:

\begin{equation}
\label{eq:krx}
D_{KRX}(\x_\ast) = {\bf k}_\ast^\top(\K\K)^{-1}{\bf k}_\ast,
\end{equation} 
where ${\bf k}_\ast=[K(\x_\ast,\x_1),\ldots,K(\x_\ast,\x_n)]^\top\in\Real^{n}$ contains the similarities between $\x_\ast$ and all points in $\X$ using $K$, and $\K\in\Real^{n\times n}$ stands for the kernel matrix containing all data similarities~\cite{Heesung05tgars}. Note that, as in the linear RX method, the KRX also requires centering the data (now in $\mathcal H$), which can be easily done\footnote{Centering in feature space can be easily done implicitly via the simple kernel matrix operation $\tilde\K\leftarrow{\bf H}{\bf K}{\bf H}$, where $H_{ij} = \delta_{ij} - \frac{1}{n}$, $\delta$ represents the Kronecker delta $\delta_{i,j}=1$ if $i=j$ and zero otherwise.}. Hereafter we assume that all kernel matrices are centered.

Note that constructing and inverting a kernel matrix of large $n$ poses a huge computational cost. A simple strategy to alleviate this problem is to draw $r$ samples randomly ($r\ll n$) and use them in the standard KRX,
which is here referred to as simple sub-sampling (SRX) and defined as
\begin{equation}
\label{eq:srx}
D_{SRX}(\x_\ast) = {\bf k}_{\ast:r}^\top(\hat{\K}\hat{\K})^{-1}{\bf k}_{\ast:r},
\end{equation} 
where $\hat\X \in\Real^{r \times d}$ is a data matrix sampled from $\X$, 
${\bf k}_{\ast:r}=[K(\x_\ast,\x_1),\ldots,K(\x_\ast,\x_r)]^\top\in\Real^{r}$ contains the similarities between $\x_{\ast}$ and $\hat{\X}$, 
and $\hat{\K}\in\Real^{r\times r}$ is a kernel matrix containing data similarities between the points in $\hat{\X}$.

\section{Efficient techniques for Kernel RX}

Kernel methods are able to fit nonlinear problems but, they do not scale well when the number of samples grow. We propose using feature map and low-rank approximation approaches to improve the efficiency of the KRX detector. We study the following approximations to the KRX method: Random Fourier features (RRX) previously studied by the authors in~\cite{RRX}, orthogonal random features (ORX), naive low-rank approximation (LRX), and Nystr\"om low-rank approximation (NRX).

\subsection{Randomized Feature Map Approaches}

\subsubsection{Random Fourier Features (RFF)}
An outstanding result in the recent kernel methods literature makes use of a classical definition in harmonic analysis to  the approximation and scalability~\cite{RFF17}. 
The Bochner's theorem states that a continuous shift-invariant kernel $K(\x,\x')=K(\x-\x')$ on $\Real^d$ is positive definite (p.d.) if and only if $K$ is the Fourier transform of a non-negative measure. 
If a shift-invariant kernel $K$ is properly scaled, its Fourier transform $p({\bf w})$ is a proper probability distribution. 
This property is used to approximate kernel functions with linear projections on a number of $D$ random features as 
$
K(\x, \x') \approx \frac{1}{D} \sum\nolimits_{i=1}^D \exp(-\imag{\bf w}_i^\top\x)\exp(\imag{\bf w}_i^\top\x'),
$
where ${\bf w}_i \in \Real^{d}$ are randomly sampled from a data-independent distribution $p({\bf w})$~\cite{RFF17}. 
Note that we can define a $2D$-dimensional {\em randomized} feature map ${\bf z}: \Real^d\to\Real^{2D}$, which can be {\em explicitly} constructed as ${\bf z}({\bf x}) = \frac{1}{\sqrt{2D}}[\cos({\bf w}_1^\top{\bf x}),\sin({\bf w}_1^\top{\bf x}),\ldots,\cos(\sin{\bf w}_D^\top{\bf x}),\sin({\bf w}_D^\top{\bf x})]^\top$ to approximate the Radial Basis Function (RBF) kernel. 

Therefore, given $n$ data points (pixels), the kernel matrix ${\bf K} \in\Real^{n\times n}$ can be approximated with the explicitly mapped data, ${\bf Z}=[{\bf z}_1\cdots{\bf z}_n]^\top\in\Real^{n\times 2D}$, and will be denoted as $\hat{\bf K}\approx{\bf Z}{\bf Z}^\top.$ 
However, we do not use such an approach in Equation~\eqref{eq:krx}, which would lead to a mere approximation with extra computational cost. 
Instead, we run the linear RX in Equation~\eqref{eq:rx_centered} with explicitly mapped points onto random Fourier features, which reduces to 
\begin{equation}
    D_{RRX} = \z_\ast^\top(\Z^\top\Z)^{-1}\z_\ast,
\end{equation}
and leads to a nonlinear randomized RX (RRX) \cite{RRX} that approximates the KRX. 
Essentially, we map the original data $\x_i$ into a nonlinear space through the explicit mapping $\z(\x_i)$ to a $2D$-dimensional space (instead of the potentially infinite feature space with $\boldsymbol{\phi}(\x_i)$), and then use the linear RX formula. This allows us to control the space and time complexity explicitly through $D$, as one has to store matrices of $n\times 2D$ and invert matrices of size $2D\times 2D$ only (see Table~\ref{spacetime_complexity}).
Typically, parameter $D$ satisfies $D\ll n$ in practical applications.

\subsubsection{Orthogonal Random Features (ORF)}
RFF has become a very practical solution for the bottleneck in kernel methods when $n$ grows. 
In RFF, frequencies $\bf{w_i}$ are sampled from a particular pdf and they act as a basis. This, however, may lead to features that are linearly dependent thus geometrically covering less space. Imposing orthogonality in the basis can be a remedy to this issue, which has led to the Orthogonal Random Features (ORF)~\cite{ORF16}. 
The linear transformation matrix of ORF is ${\bf W}_{\text{ORF}}=\frac{1}{\sigma}{\bf S}{\bf Q}$, where ${\bf Q}$ is a uniformly distributed random orthogonal matrix. 
The set of rows of $Q$ forms a basis in $\mathbbm{R}^d$. ${\bf S}$ is a diagonal matrix, with diagonal entries sampled i.i.d. from the $\chi$-distribution with $d$ degrees of freedom. ${\bf S}$ makes the norms of the rows of ${\bf SQ}$ and ${\bf W}$ (with all the frequencies of RFF) identically distributed.
Theoretical results show that ORF achieves lower error than RFF for the RBF kernel~\cite{ORF16}. This approach follows the above RFF philosophy, and the final anomaly score is now: 
\begin{equation}
    D_{ORX} = \z_\ast^\top(\Z^\top\Z)^{-1}\z_\ast,
\end{equation}
where each frequency ${\bf w}_i$ is a row of $\bf{W}_{\text{ORF}}$ and ${\bf Z}$ is the matrix formed by the mappings ${\bf z}({\bf x}_i)$ of each element in the dataset, and ${\bf z_\ast}$ is the mapping of a pixel to be tested.

\subsubsection{Nystr\"om Approximation}
The Nystr\"om method selects a subset of samples to construct a low-rank approximation of the kernel matrix~\cite{Nystrom}. This method approximates the kernel function as $K(\x_*, \x) \approx {\bf k}_{*:r}^\top{\hat\K}^{-1} {\bf k}_{\x:r}$,
where ${\bf k}_{\x:r}$ contains the similarities between $\x$ and all $r$ points, and $\hat{\K}\in\Real^{r\times r}$ stands for the kernel matrix between the points in $\hat\X$. Therefore, ${\bf k}_\ast$ can be expressed as:
\begin{equation}
 \label{eq:kstar_approximation}
 {\bf k}_\ast \approx  {\R}^\top {\hat\K}^{-1} {\bf k}_{\ast:r},
\end{equation} 
where ${\R}\in\Real^{r \times n}$ is a matrix which contains similarities between the points in $\hat\X$ and the points in $\X$. The similarities were computed using the standard RBF kernel function $k(\x,\y)=\exp(-\|\x-\y\|^2/(2\sigma^2))$.

Using the above definition given in~\eqref{eq:kstar_approximation}, the Nystr\"om method approximates the kernel matrix $\K$
\begin{equation}
 \label{eq:K_approximation}
 \K \approx {\R}^\top {\hat\K}^{-1} {\R}.
\end{equation}
by plugging~\eqref{eq:kstar_approximation} and \eqref{eq:K_approximation} into~\eqref{eq:krx}, one can define the low-rank approximation of KRX:
\begin{equation}
\label{eq:nrx2}
D_{NRX}(\x_\ast) = 
{\bf k}_{\ast:r}^\top {\hat\K}^{-1} {\R}
({\R}^\top \M {\R})^{-1} 
{\R}^\top {\hat\K}^{-1} {\bf k}_{\ast:r},
\end{equation} 
where $\M = {\hat\K}^{-1} {\R} {\R}^\top {\hat\K}^{-1}$ while $\M \in\Real^{r \times r}$. 
Since $\R$ is not a squared matrix $(r<n)$, it is rank deficient, and we propose to use the pseudoinverse instead of the inverse of ${\R}^\top \M {\R}$. By doing this, most of the terms cancel, leading to a more compact equation for the NRX:

\begin{equation}
\label{eq:nrx}
D_{NRX}(\x_\ast) = {\bf k}_{\ast:r}^\top (\R\R^{\top})^{\dagger} {\bf k}_{\ast:r}.
\end{equation}

Note that NRX involves the inversion of an $r\times r$ matrix which is much more efficient compared to KRX.
In addition, the Nyström approach is more generic than using random Fourier feature approaches, as it allows one to approximate all positive semidefinite kernels, not just shift-invariant kernels. Furthermore, this approximation is data-dependent (i.e. the basis functions are a subset of estimation data itself) which could translate into better results~\cite{NystromVSRandomFeatures}.

\subsubsection{Connection to reduced-set methods} Reduced-set techniques were successfully used to obtain sparse kernel methods and low rank approximations of multivariate kernel methods~\cite{Arenas2013}. This methodology can be applied to approximate KRX which leads to  equation~\eqref{eq:nrx}. In this approach, the data matrix $\X\in\Real^{n\times d}$ is subsampled into $\hat{\X} \in\Real^{r \times d}$, $r\ll n$, and mapped into $\hat{\boldsymbol{\Phi}}\in\Real^{r \times d_{\mathcal H}}$, which, by using \eqref{eq:krx_Hilbert}, we obtain the LRX formula:
\begin{equation}
\label{eq:krx_Hilbert_lowRank}
D_{LRX}(\x_\ast) = {\boldsymbol{\phi}(\x_\ast)}^\top {\hat{\boldsymbol\Phi}}^\top
({\hat{\boldsymbol\Phi}} \boldsymbol{\Phi}^\top {\boldsymbol\Phi} {\hat{\boldsymbol\Phi}}^\top)^{-1}
{\hat{\boldsymbol\Phi}} \boldsymbol{\phi}(\x_\ast).
\end{equation} 
Identifying ${\bf k}_{\ast:r}={\hat{\boldsymbol\Phi}} \boldsymbol{\phi}(\x_\ast)$ and $\R = {\hat{\boldsymbol\Phi}} \boldsymbol{\Phi}^\top$,~\eqref{eq:krx_Hilbert_lowRank} leads to:
\begin{equation}
\label{eq:lrx} 
D_{LRX}(\x_\ast) = {\bf k}_{\ast:r}^\top (\R\R^{\top})^{-1} {\bf k}_{\ast:r},
\end{equation} 
which just differs from~\eqref{eq:nrx} in the inverse of $\R\R^{\top}$, and when $\R$ is full rank they are the same. In the following and in the experiments, we will use only $D_{NRX}$ instead of $D_{LRX}$ as both are mathematically equivalent.

\subsection{Space and Time Complexity}
Table~\ref{spacetime_complexity} gives the theoretical computational complexity of the benchmark methods (RX, KRX, SRX) and proposed methods (RRX, ORX, NRX) presented in this paper. In this study, we assume $d<D<r\ll n$ since we aim to deal with big data settings. Besides, KRX becomes sufficiently efficient when $n$ is small, e.g. $n < 4000$ for a $200 \times 200$ image.
As seen in Table~\ref{spacetime_complexity}, RX provides the best efficiency; thus, it should be employed for scenes where the data is Gaussian distributed.
However, KRX and the proposed KRX approximations should be used for nonlinear distributions. 
Clearly, KRX is the least efficient compared to the proposed approximations, and it is also not applicable to big data. 
Feature map methods, e.g. RRX and ORX, provide the best computational efficiency for nonlinear (i.e non-Gaussian) distributions, while low-rank approximation methods, e.g. LRX and NRX, are also efficient yet relatively slower compared to the feature map methods.
Thus, one should choose the proper method based on the image distribution characteristics~\cite{Manolakis07a,Keshava04}, detection performance requirements, and computational resource limitations. 
These conclusions are assessed experimentally in the following section.
\begin{table}[!ht]
\begin{center}
\captionsetup{skip=1pt}
\caption{Memory and time complexity for all methods.}
\label{spacetime_complexity}
\begin{tabular}{|l|l|l|| l|l|l|l|}
\hline
\rowcolor[gray]{.60}
& \multicolumn{2}{c||}{Space}     & \multicolumn{4}{c}{Time}       \\ \hline
\rowcolor[gray]{.90}
Method & $T$     & $C^{-1}$   & $T$      & $C$      & $C^{-1}$   & $AD$       \\ \hline \hline
RX     & $-$     & $d^2$      & $-$      & $nd^2$   & $d^3$      & $nd^2$     \\ \hline
RRX \& ORX   & $nD$    & $D^2$      & $ndD$    & $nD^2$   & $D^3$      & $nD^2$     \\ \hline
NRX   & $nr$    & $r^2$      & $ndr$    & $nr^2$    & $r^3$      & $nr^2$     \\ \hline
KRX    & $n^2$   & $n^2$      & $n^2d$   & $n^3$    & $n^3$      & $n^3$      \\ \hline
\end{tabular}
\vspace*{1pt}
\newline\footnotesize{$T$ is transformation of image into a nonlinear space.}
\newline\footnotesize{$C$ is matrix (covariance, kernel etc.) and $C^{-1}$ is its inverse.}
\end{center}
\end{table}

\section{Experimental Results} 
\label{sec:data}
\begin{figure*}[t]
\centering 

\subfloat[]{\includegraphics[width= 0.225\textwidth]{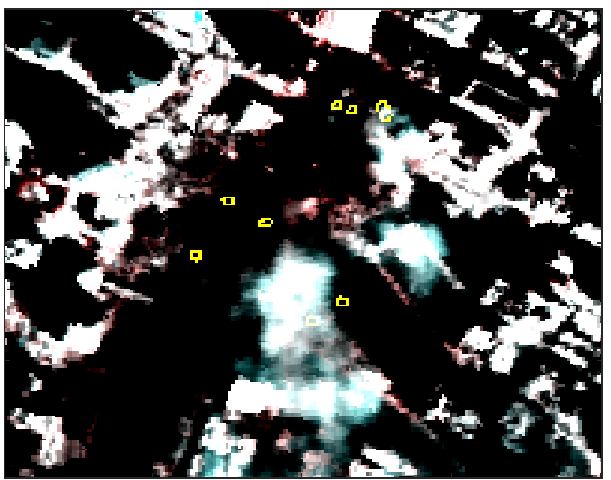}} \hspace{1mm}
\subfloat[]{\includegraphics[width= 0.225\textwidth]{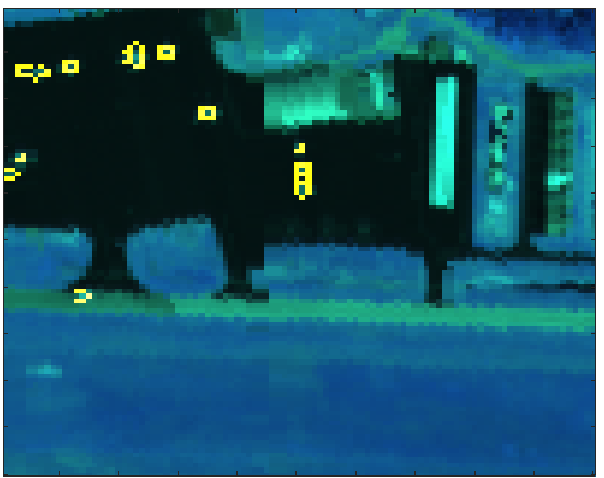}} \hspace{1mm}
\subfloat[]{\includegraphics[width= 0.225\textwidth]{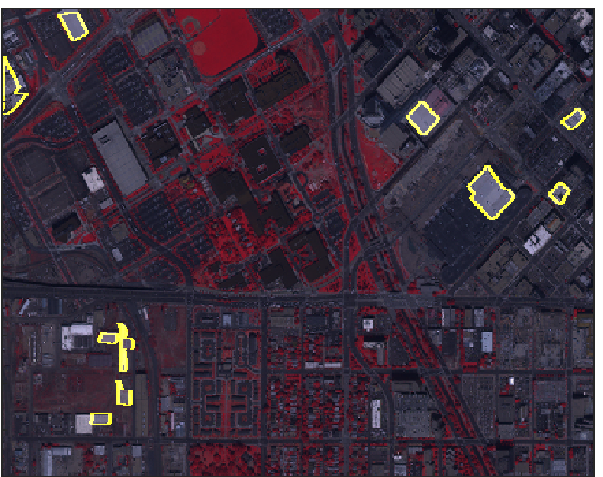}}\hspace{1mm}
\subfloat[]{\includegraphics[width= 0.225\textwidth]{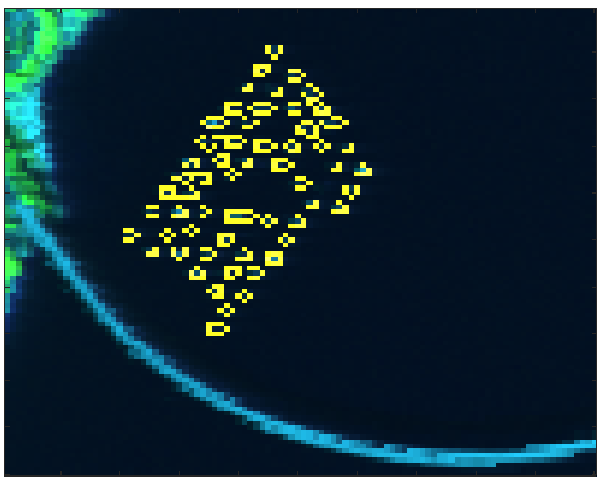}} \hspace{1mm}

\caption{\small Images with anomalies (outlined in yellow) in four scenarios: (a) consequences of the hot spots corresponding to latent fires at the World Trade Center (WTC) in NYC (extension of anomalous pixels represents the 0.23\% of the image), (b) urban area where anomalies are vehicles in Gainesville city (0.52\%), (c) Quickbird multispectral images acquired over Denver, the anomalies are roofs in an urbanized area (1.6\%), and (d) a beach scene where the anomalies are ships captured by AVIRIS sensor (2.02\%) over San Diego, USA.}\label{fig:rgb2}
\end{figure*}
This section analyzes the performance of the proposed nonlinear RX anomaly detection methods.
We performed tests in four real examples, and tested robustness using the area under curve (AUC) of receiver operating characteristic (ROC) curves. We provide illustrative source code for all methods in \href{http://isp.uv.es/code/fastrx.html}{http://isp.uv.es/code/fastrx.html}
\subsection{Data collection and experimental setup}

We collected multispectral and hyperspectral images acquired by the Quickbird and AVIRIS sensors. Fig.~\ref{fig:rgb2} showcases the scenes used in the experiments. The AD scenarios consider anomalies related to: latent fires, vehicles, urbanization (roofs) and ships~\cite{Guo2016,ABU,jose}. Table~\ref{table:database} summaries relevant attributes of the datasets such as sensors, spatial and spectral resolution.

Parameter estimation is required for the RX, KRX, RRX, ORX and NRX. 
First of all, the KRX method and its proposed variants involve the optimization of the $\sigma$ parameter of the RBF kernel.
For the feature map approaches (RRX and ORX), the number of basis, $D$, parameter should be optimized.
Whereas, for low-rank approximations (NRX), the number of random sub-samples, $r$, parameter should be optimized. 

\begin{table}[!]
\centering
\caption{Images attributes used in the experimentation dataset. }\label{table:database}
\label{ROC_table}
\begin{tabular}{ |l|l|c|l|l|} 
\hline\hline
\rowcolor[gray]{.90} \bf{Images} & \bf{Sensor}&\bf{Size} & \bf{Bands}  & \bf{Resolution}\\ \hline\hline
WTC   & AVIRIS  & 200 x 200  &  224  & 1.7 m 
\\\hline
Gainesville & AVIRIS    & 100 x 100  &  190 & 3.5 m
\\\hline
Denver  & Quickbird & 500 x 684     &   4   & 1m-4m          
\\\hline
 San Diego   & AVIRIS      & 100 x 100  &  193  & 7.5 m
\\\hline            

\hline

\end{tabular}
\end{table}

We adopted a cross-validation scheme to select all the involved parameters: number of Fourier basis $D$, rank $r$, and RBF parameter $\sigma$. We selected parameters using different data sizes ranging between $10^3$ and $3\times10^4$ samples.

\subsection{Numerical comparison}

\begin{figure}[!htp]
    \centering
    \begin{tabular}{cc}
    WTC & Gainesville \\ 
    
    \includegraphics[width=0.47\columnwidth]{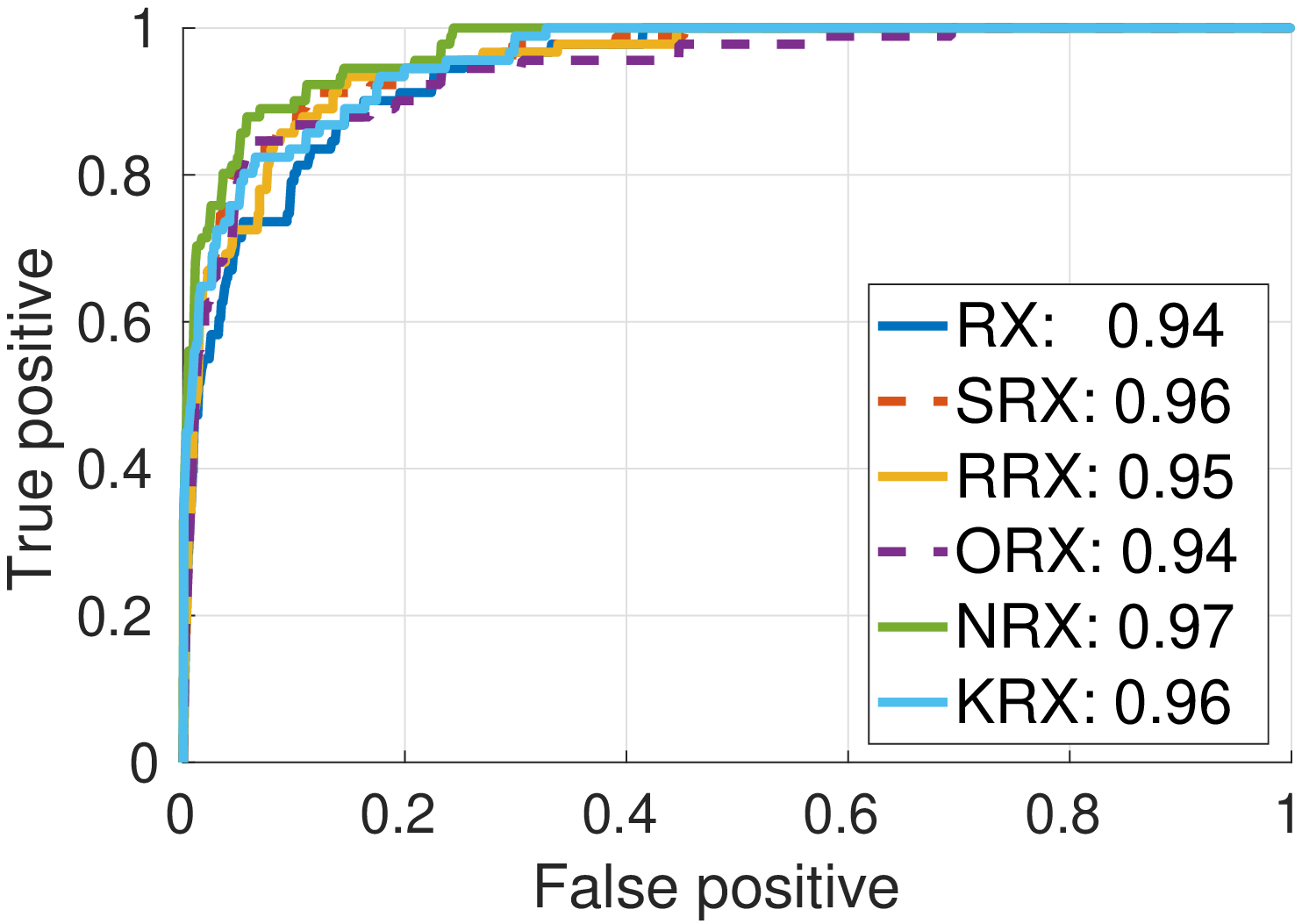} &
    \includegraphics[width=0.47\columnwidth]{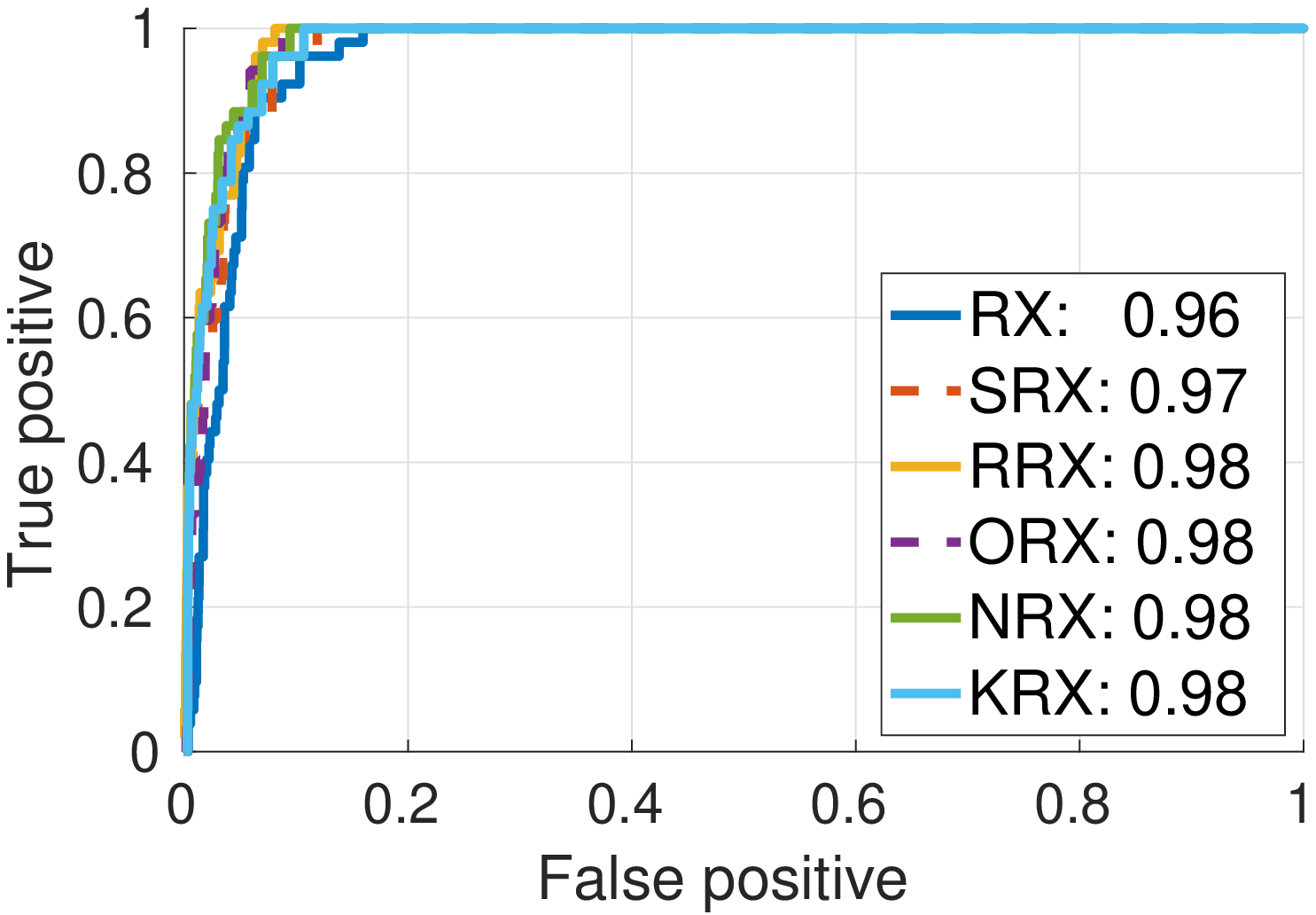}\\
    
    Denver & San Diego\\
    
    \includegraphics[width=0.47\columnwidth]{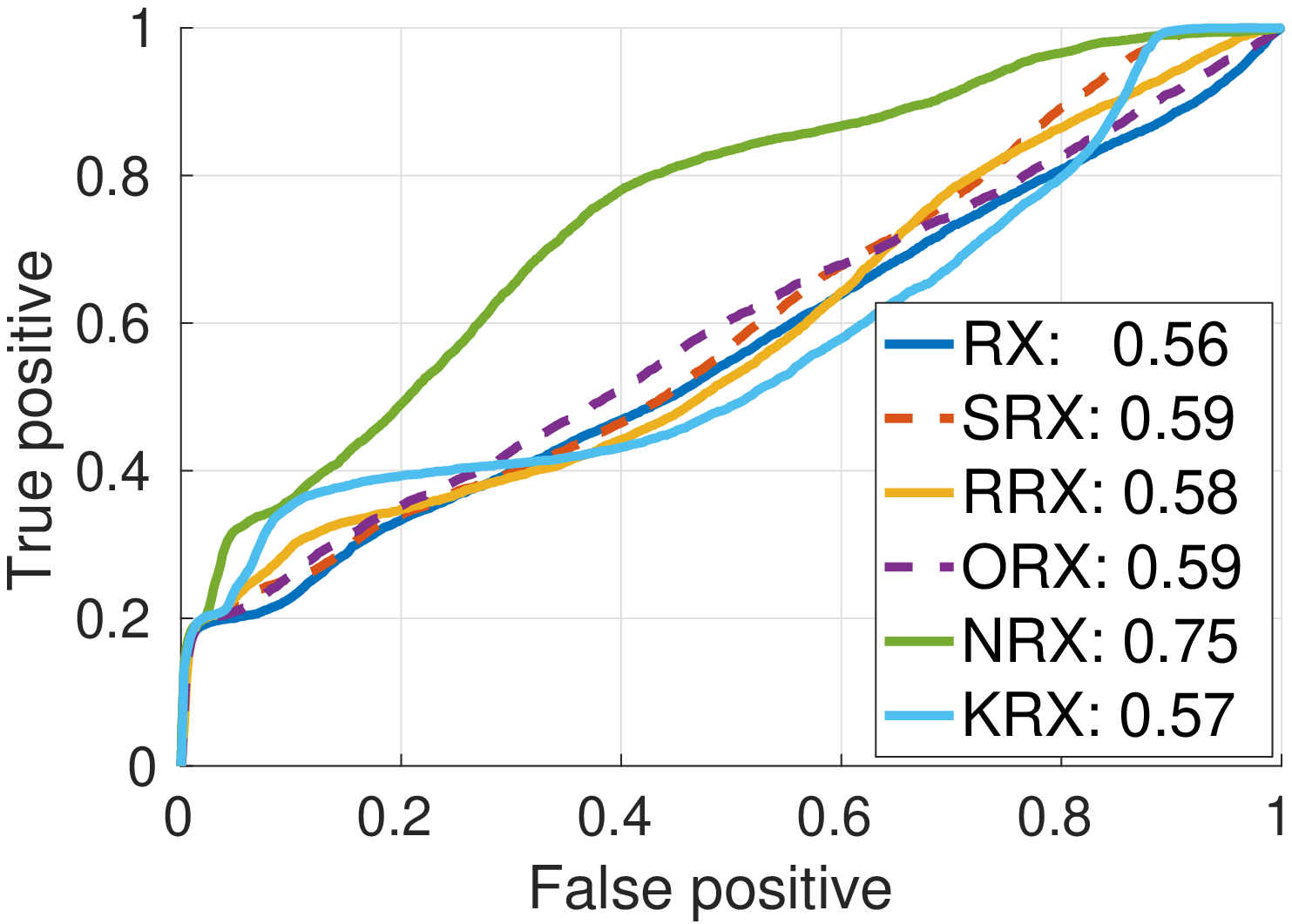}&
    \includegraphics[width=0.47\columnwidth]{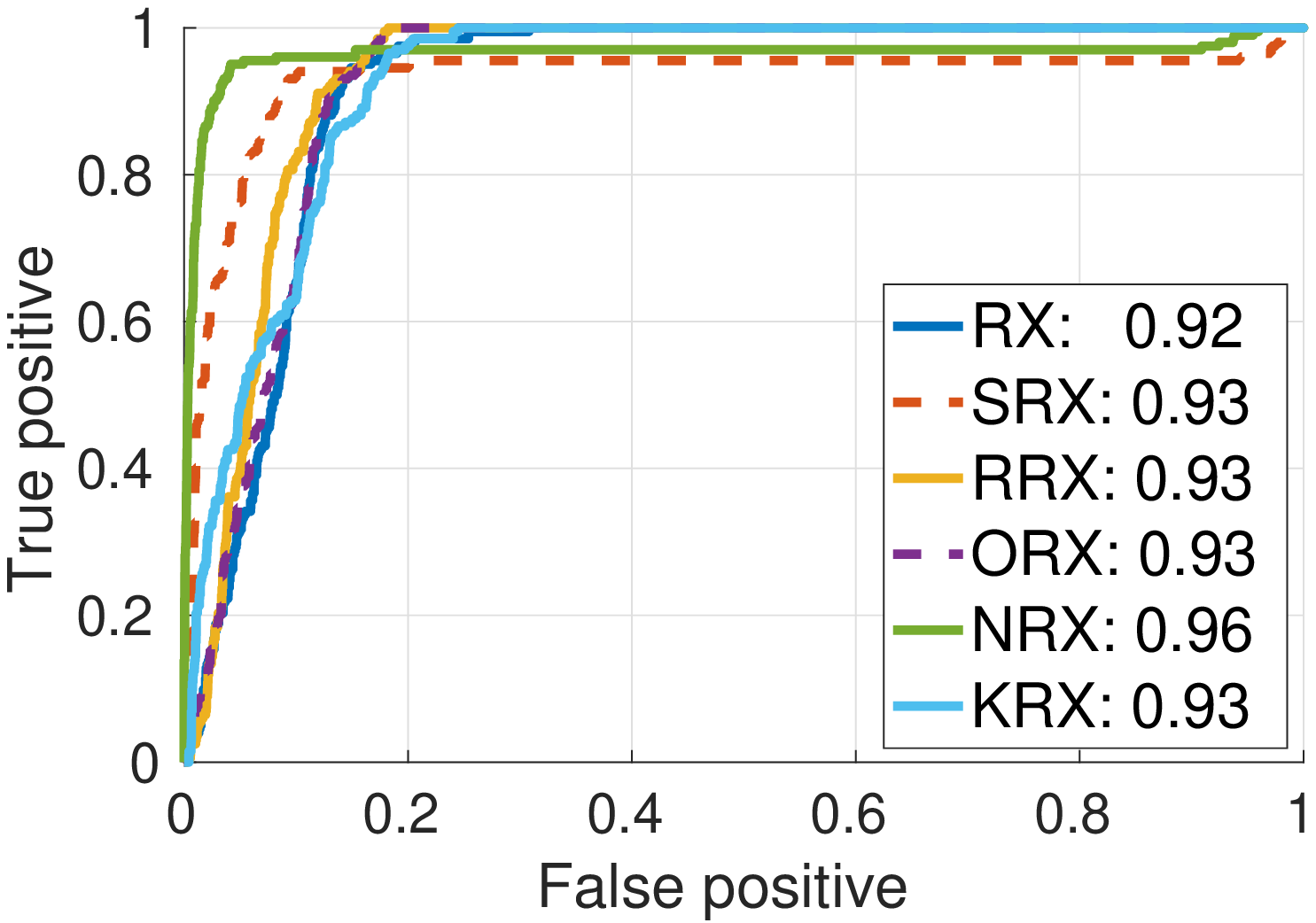}
    \end{tabular}
    \caption{ROC curves in linear scale for all scenes. Numbers in legend display the AUC values for each method. }
    \label{fig:ROC_log}
\end{figure}

We report the averaged AUC results for all cases with $1000$ runs (standard deviations were always lower than $3\times 10^{-3}$ and hence are not reported). Figure~\ref{fig:ROC_log} shows that nonlinear methods improve detection over the linear RX and NRX outperforms the other approximations in three out of the four images. The AUC values of KRX are related to the inversion of a relatively big matrix. This raises the issues of poorly estimated matrices (with a huge condition number) which are also computationally expensive to invert ($\mathcal{O}(n^3)$).
However, all the proposed fast kernel RX methods have the advantage of solving both issues. Firstly, thanks to the cross-validation procedure, an estimate of the optimal number of features (RRX, ORX) or samples (NRX) can be obtained, allowing to better capture the intrinsic dimensionality of the mapped data. In a previous work~\cite{Morales18}, we showed that optimizing the number of frequencies in random Fourier features approaches acts as an efficient regularizer leading to better estimates with a reduced number of frequencies needed. And secondly, fast versions are able to obtain better performance in AUC metric at a fraction of the cost (see Fig.~\ref{fig:ROC_log}).

\begin{figure}[!htp]
    \centering
   
    \includegraphics[width=0.48\columnwidth]{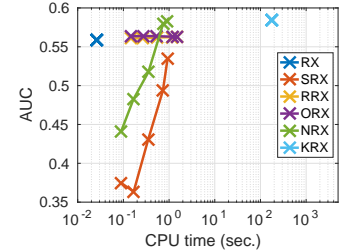} 
    
    \caption{ CPU execution time versus the AUC values for $n=3000$ pixels, crosses corresponds to different rank  values for Denver image.}
    \label{fig:training_time}
\end{figure}

\subsection{On the computational efficiency}


\begin{figure*}[!t]
    \centering
    \begin{tabular}{cccc}
        WTC & Gainesville & Denver & San Diego\\
    \includegraphics[width=3.7cm,height=3.7cm]{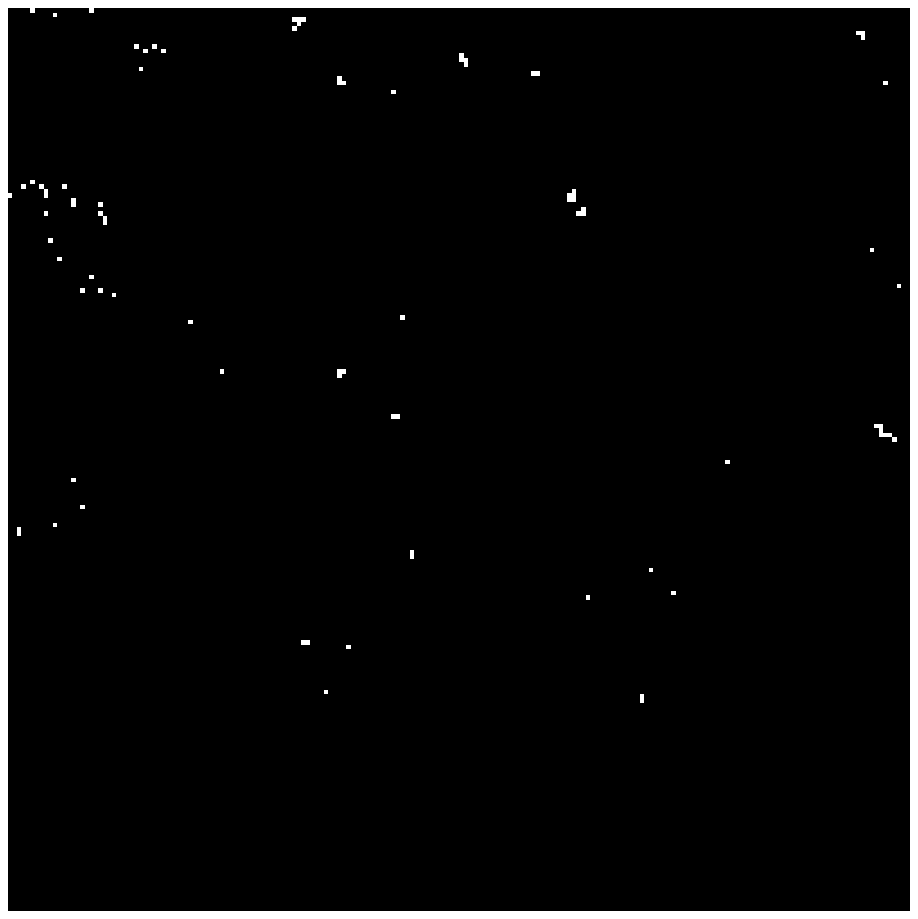}\hspace{1mm} &
    \includegraphics[width=3.7cm,height=3.7cm]{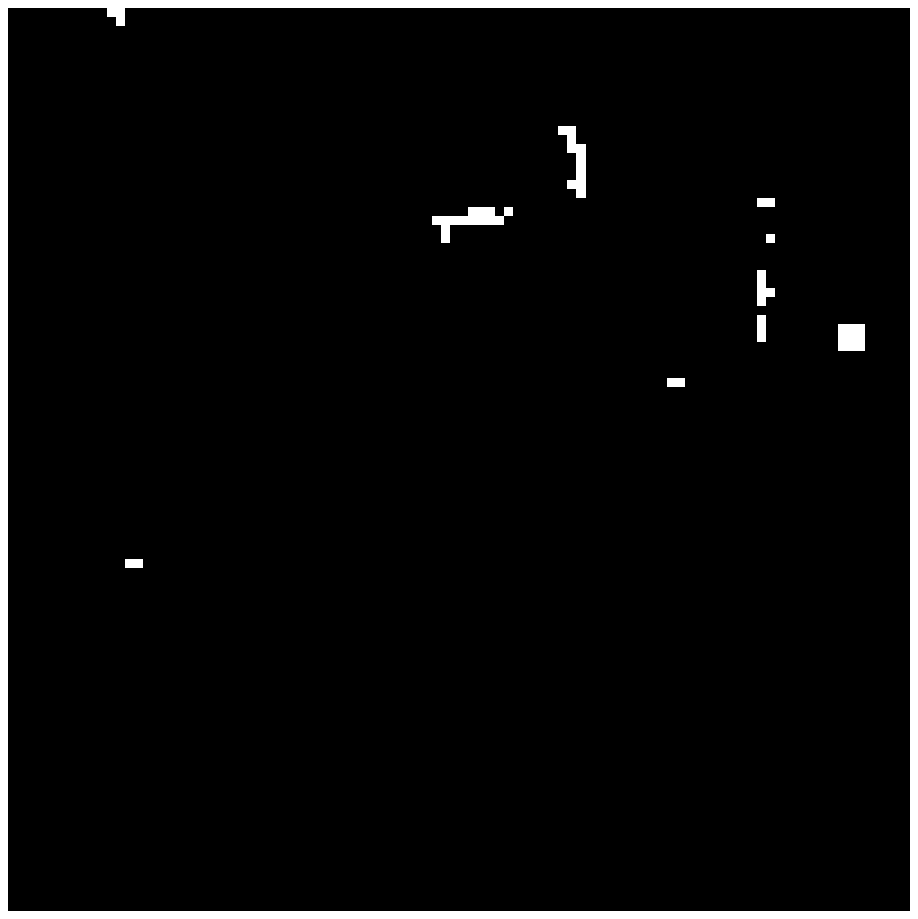}\hspace{1mm}&
    \includegraphics[width=3.7cm,height=3.7cm]{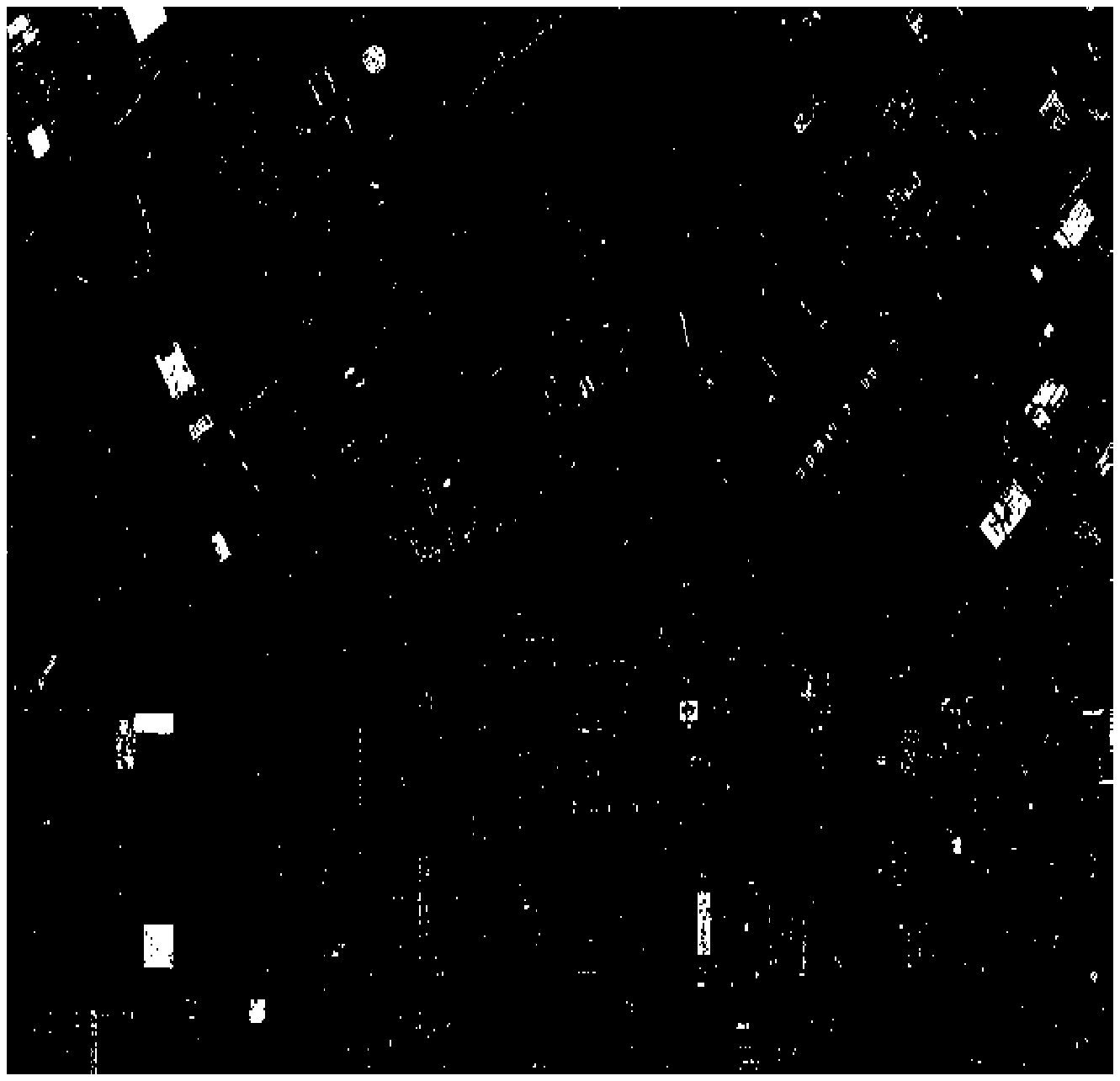}\hspace{1mm}&
    \includegraphics[width=3.7cm,height=3.7cm]{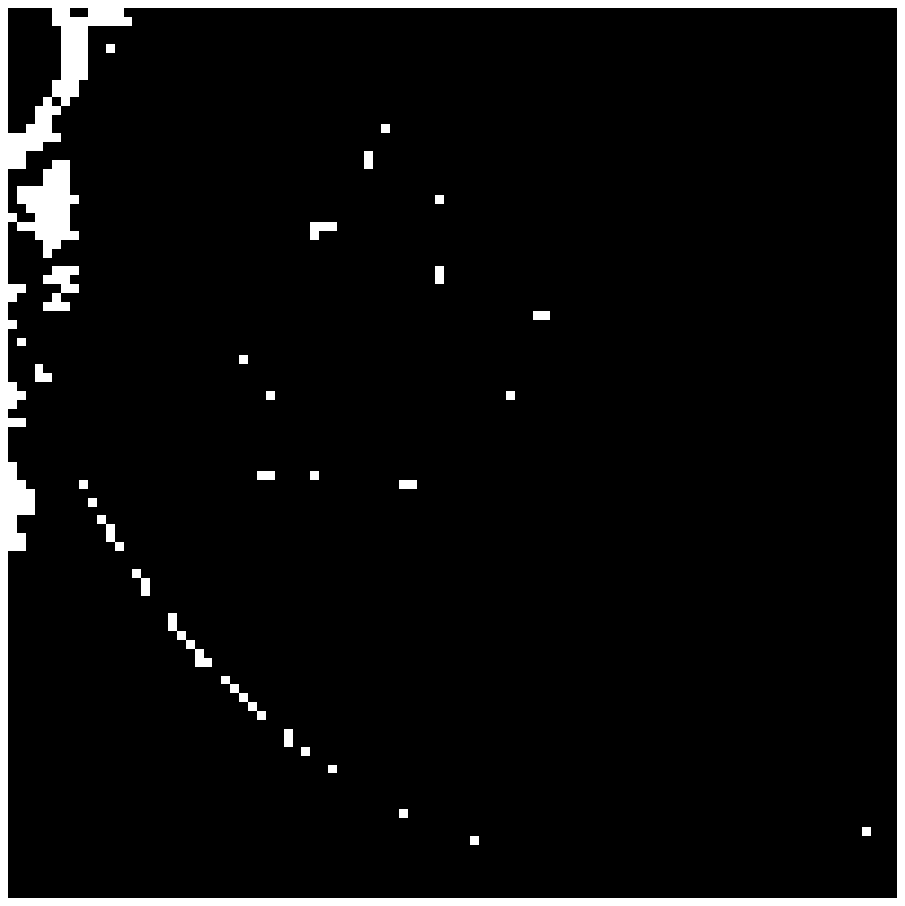}\\
    RX (0.94) & RX (0.96) & RX (0.56) & RX (0.92) \\
    \includegraphics[width=3.7cm,height=3.7cm]{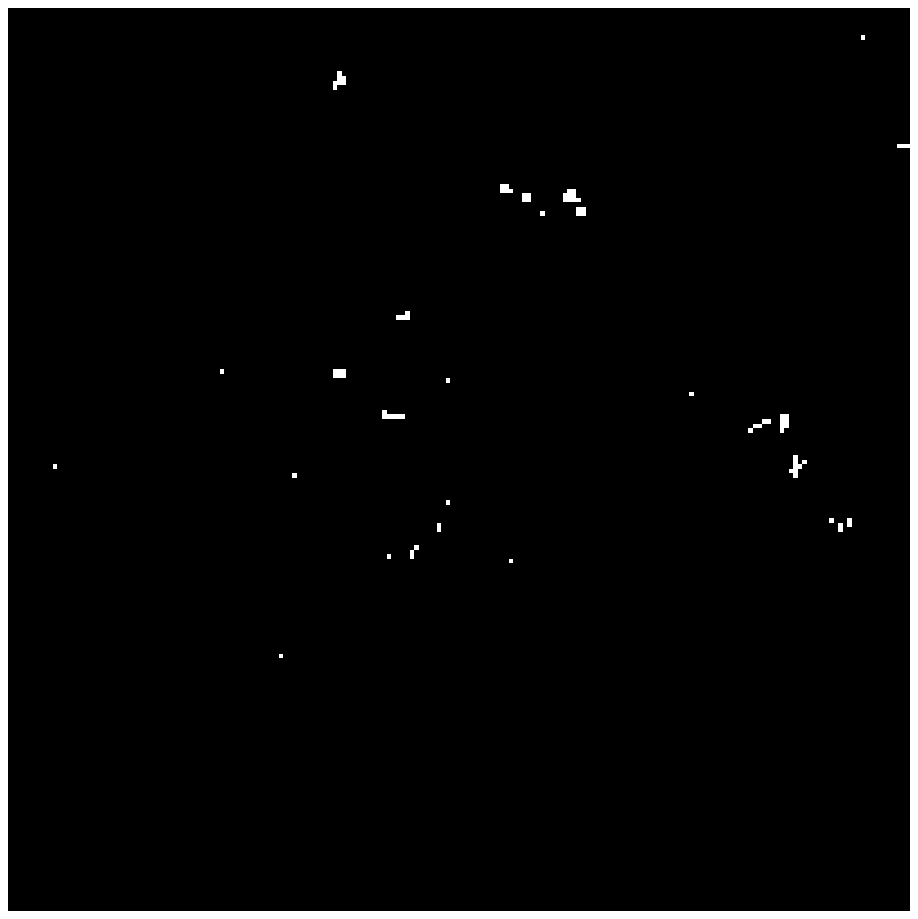}\hspace{1mm} &
    \includegraphics[width=3.7cm,height=3.7cm]{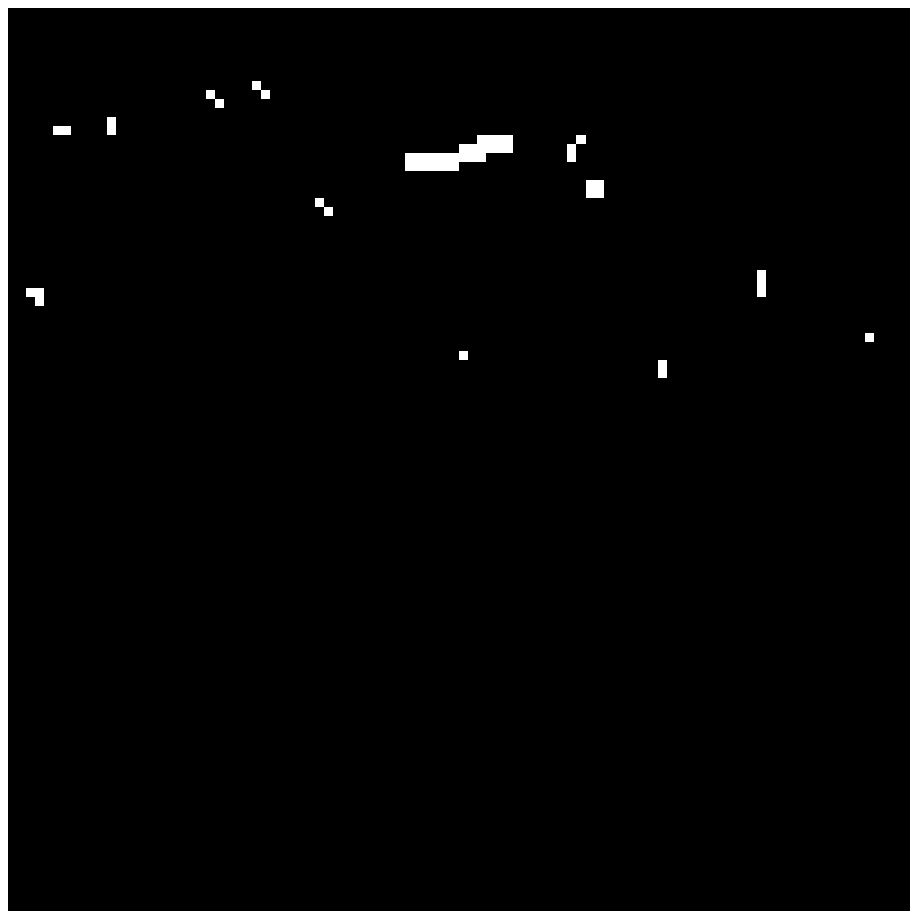}\hspace{1mm}&
    \includegraphics[width=3.7cm,height=3.7cm]{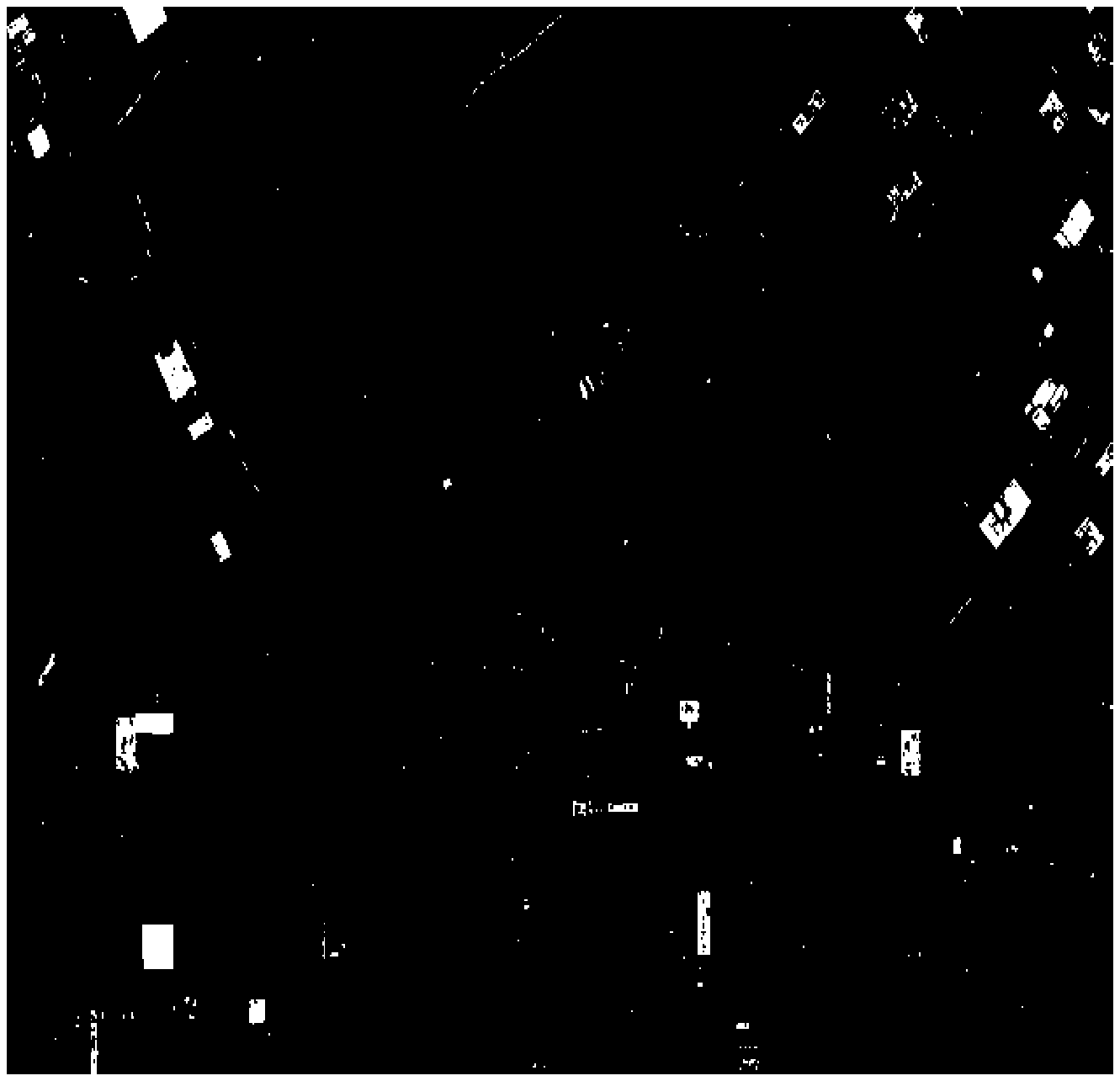}\hspace{1mm}&
    \includegraphics[width=3.7cm,height=3.7cm]{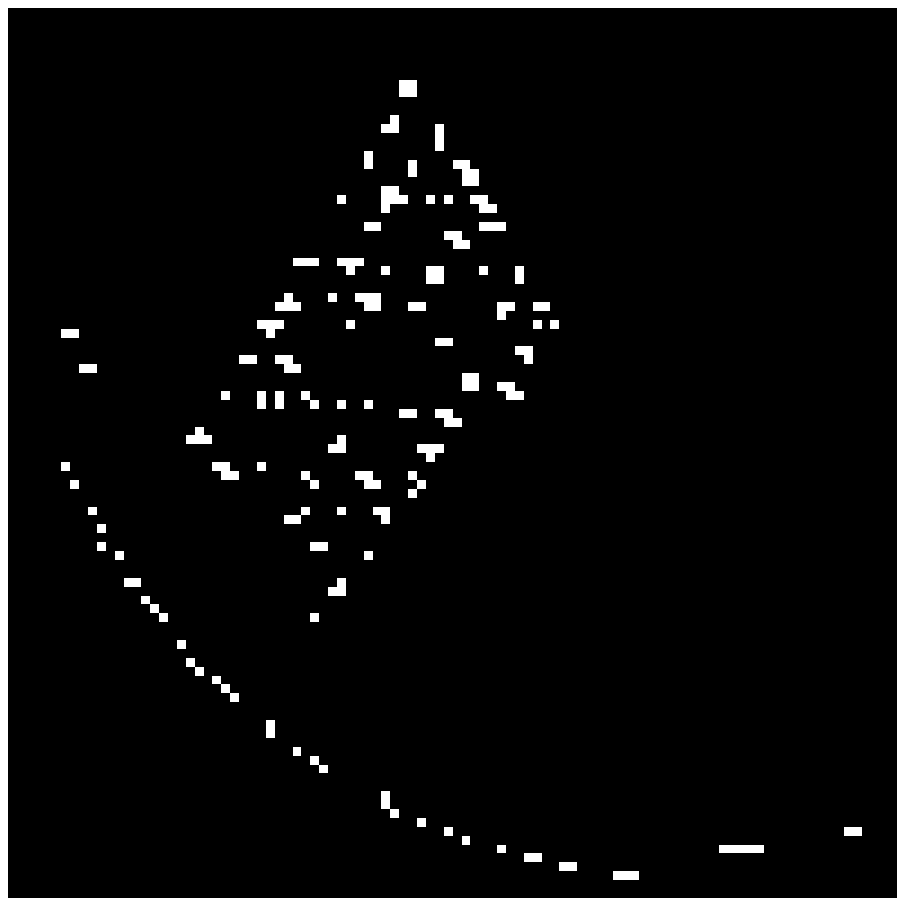}\\
    NRX (0.97) & NRX (0.98) & NRX (0.75) & NRX (0.96)
    \end{tabular}
    \caption{Anomaly detection maps for best thresholds (top: the best linear RX (AUC) results, bottom: best nonlinear RX (AUC) method).}
    \label{fig:maps}
\end{figure*}

Figure~\ref{fig:training_time} illustrates the trade-off between the computational execution time and the AUC. The crosses indicate different values of rank ($D$ or $r$ parameters) in the set $\{50, 100, 200, 400, 500\}$ and the number of pixels was fixed to $n=3000$. The optimal parameters estimated for KRX are used for the fast approaches. KRX has the best AUC values in all the images. NRX and SRX are more sensitive to rank values. RRX and ORX are almost insensitive to the rank but results do not improve when the rank increases, thus limiting their performance. The combination of lower spectral information and the ambiguity of the class (note that the anomaly class `urbanized' can be confused with a pervasive class `urban') makes the Quickbird scene a very difficult problem (lower AUCs). In this situation, as the rank parameter $r$ for the NRX method grows, it approximates the KRX algorithm.
 In Figure~\ref{fig:maps}, the RX detector (top row) is shown against the best detector obtained (bottom row). The best result in AUC was achieved by the NRX in all the images. It is worth mentioning the good results in detection achieved by the NRX in all the scenes, which can be visually compared the linear version.

\section{Conclusions} 
\label{sec:conclusions}

In this letter, we introduced a family of efficient nonlinear anomaly detection algorithms based on the RX method. We used the theory of reproducing kernels, and proposed several efficient methods. 
The kernel Reed-Xiaoli (KRX) detector was improved using efficient and fast techniques based on feature maps and low-rank approximations.
Among all methods, both the Nystr\"om and the equivalent low-rank (LRX) approximation achieves the best results and yields a more efficient and accurate non-linear RX method to be applied in practice. For future research, we plan to study the behaviour of fast approximations for alternative KRX variants~\cite{Theiler_GRSL,Theiler_WHISPERS}. Note that the presented methodologies for fast KRX can be applicable to other kernel anomaly detectors, in local settings, and for real-time detection.

\bibliographystyle{IEEEtran}
\bibliography{bibliography}

\end{document}